\documentclass[11pt,a4paper]{article}
\usepackage[hyperref]{acl2019}
\usepackage{amsmath}
\usepackage{booktabs}
\usepackage{graphicx}
\usepackage{enumitem}
\usepackage{times}
\usepackage{latexsym}

\usepackage{url}

\aclfinalcopy 

\title{Big Bidirectional Insertion Representations for Documents}

\author{Lala Li \\
  Google Research, Brain Team \\
  \texttt{lala@google.com} \\\And
  William Chan \\
  Google Research, Brain Team \\
  \texttt{williamchan@google.com} \\}

\date{}

\begin{document}
\maketitle
\begin{abstract}
  The Insertion Transformer is well suited for long form text generation due to its parallel generation capabilities, requiring $O(\log_2 n)$ generation steps to generate $n$ tokens. However, modeling long sequences is difficult, as there is more ambiguity captured in the attention mechanism. This work proposes the Big Bidirectional Insertion Representations for Documents (Big BIRD), an insertion-based model for document-level translation tasks. We scale up the insertion-based models to long form documents. Our key contribution is introducing sentence alignment via sentence-positional embeddings between the source and target document. We show an improvement of +4.3 BLEU on the WMT'19 English$\rightarrow$German document-level translation task compared with the Insertion Transformer baseline.
\end{abstract}

\section{Introduction}

Recently, insertion-based models  \cite{stern-icml-2019,welleck-icml-2019,gu-arxiv-2019,chan-arxiv-2019} have been introduced for text generation. Unlike traditional autoregressive left-to-right models \cite{cho-emnlp-2014,sutskever-nips-2014,vaswani-nips-2017}, insertion-based models are not restricted to generating text sequences in a serial left-to-right manner, but these models are endowed with the capabilities of parallel generation. More specifically, \citet{stern-icml-2019,chan-arxiv-2019} showed that we can teach neural nets to generate text to follow a balanced binary tree order. An autoregressive left-to-right model would require $O(n)$ generation steps to generate $n$ tokens, whereas the Insertion Transformer \cite{stern-icml-2019} and KERMIT \cite{chan-arxiv-2019} following a balanced binary tree policy requires only $O(\log_2 n)$ generation steps to generate $n$ tokens. This is especially important for long-form text generation, for example, Document-Level Machine Translation.

Document-Level Machine Translation is becoming an increasingly important task. Recent research suggests we are nearing human-level parity for sentence-level translation in certain domains \cite{hassan-arxiv-2018}, however, we lag significantly behind in document-level translation \cite{laubli-emnlp-2018}. Various papers have proposed incorporating context for document-level translation \cite{dowmunt-wmt-2019}, which has been shown to improve translation quality. There are two primary methods to include context in a document-level machine translation model compared to a sentence-level translation model.
\begin{enumerate}[itemsep=0pt]
    \item \textbf{Source Contextualization.} We can include source context, wherein when we generate the target sentence, we can condition on the corresponding source sentence and its neighbours, or even the whole source document. This allows the target sentence to be contextualized to the source document.
    \item \textbf{Target Contextualization.} We can include target context, wherein when we generate the target sentence, we can condition on all the target tokens generated thus far in the whole document. This allows the target sentence to be contextualized to other target sentences.
\end{enumerate}

Target contextualization is especially difficult in an autoregressive left-to-right model (i.e., Transformer \cite{vaswani-nips-2017}), the model must generate the whole document in linear fashion, which would be prohibitively expensive costing $O(n)$ iterations to generate $n$ tokens. Additionally, the model is unable to model bidirectional context, since the text is always generated in a left-to-right manner.
Some prior work have focused on utilizing block coordinate descent like algorithms during inference \cite{maruf-acl-2018}, however this adds complexity and additional runtime cost during inference.

\begin{figure*}[t]
\centering
\includegraphics[trim=28 155 125 25,clip,width=\textwidth]{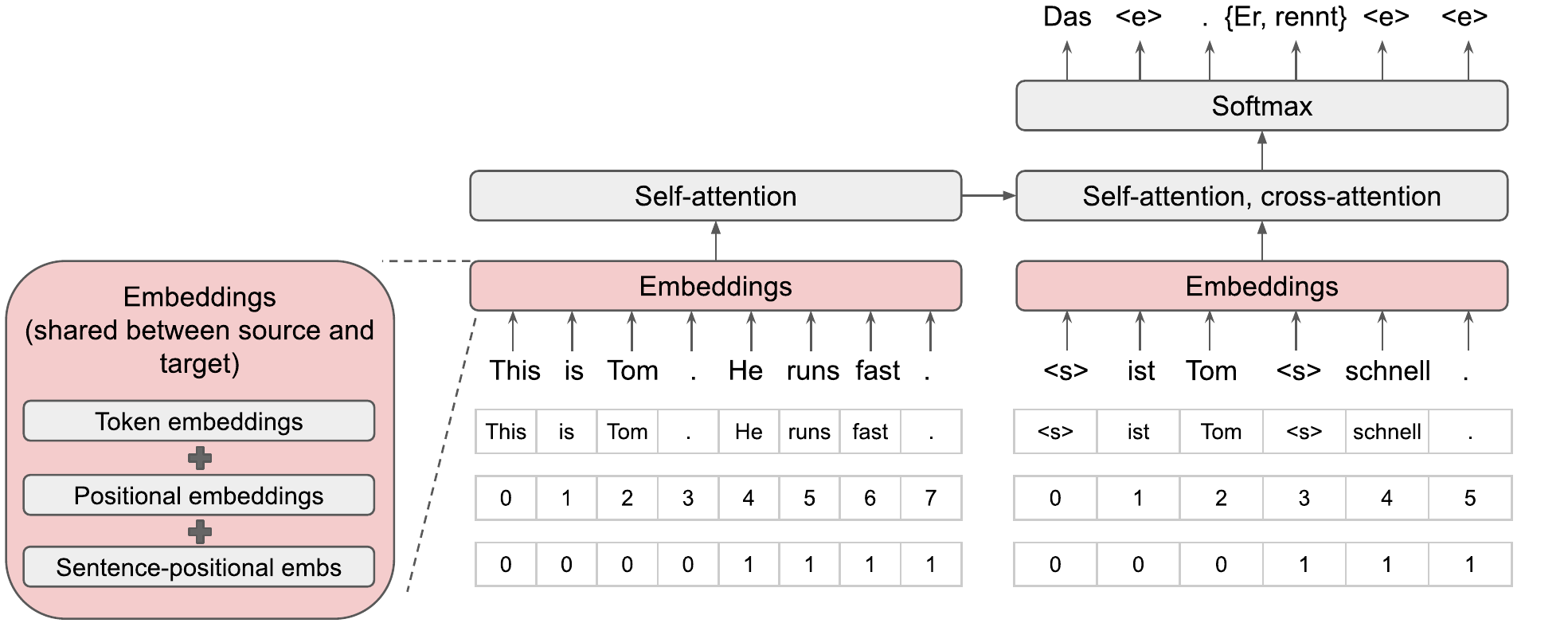}
\caption{Big Bidirectional Insertion Representations for Documents}
\label{fig:model}
\end{figure*}

Insertion-based models, for example, the Insertion Transformer \cite{stern-icml-2019} is one potential solution. The Insertion Transformer can generate text following a balanced binary tree order. It requires $O(\log_2 n)$ iterations to generate $n$ tokens, offering significant inference time advantages over a serial generation model. The source document is naturally fully conditioned on, which provides full source contextualization. Additionally, the generation order offers bidirectional contextualization, permitting target contextualization that is not solely on a left-to-right basis.

In this paper, we present Big Bidirectional Insertion Representations for Documents (Big BIRD). We address the limitations of scaling up the Insertion Transformer to document-level machine translation. We present a model that can model long-form documents with thousands of tokens in a fully contextualized manner.

\section{Big BIRD}
\label{sec:bigbird}
In this section, we present Big Bidirectional Representations for Documents (Big BIRD). Big BIRD is an extension of the Insertion Transformer \cite{stern-icml-2019}, scaling up from sentences to documents. The key contributions are 1) extending the context window size to cover a document, and 2) informing the model of sentence positional information, which are aligned between source and target sentences.

\textbf{Insertion Transformer.} In the Insertion Transformer \cite{stern-icml-2019}, sequences are generated via insertion operations. In the context of Machine Translation, there is a source canvas $x$ and a target canvas $y$, where the target canvas is updated at each iteration via inserting one token at each plausible location. At time $t$ during training, a hypothesis target canvas $\hat{y}_t$ must be a subsequence of the final output. For example, if the final output is $[A,B,C,D,E]$, then $\hat{y}_t = [B,D]$ would be a valid intermediate canvas, in which case the model would be taught to predict $[A,C,E]$. The model is taught to insert multiple tokens at incomplete slots, or predict end-of-slot for completed slots. The intermediate canvases are uniformly sampled from the ground truth target sequence. During inference, the target canvas starts empty, and tokens will be inserted iteratively until the model predicts to insert empty tokens everywhere, or the sequence has exceeded the specified maximum length.

\textbf{Larger Context and Sentence-Positional Embeddings.} Longer sequences lead to more uncertainties for the Insertion Transformer. For example, if a token in $\hat{y}_t$ appears in multiple sentences in the final output, there is ambiguity to the model which sentence it belongs to (and therefore where to attend to on both the source and target canvases). While there is location information endowed in the Transformer model, we hypothesize that token level positional information is insufficient (especially since we have limited training data). We believe that endowing the model with sentence-level positional information (i.e., which sentence each token belongs to) may help significantly disambiguate in such situations and help the model build a more robust attention mechanism.

Based on this motivation and assuming that the datasets have not only parallel documents, but also sentence alignment between source and target documents (which is true for WMT'19 document-level translation), we use sentence-positional embeddings on both the source and target sequences as shown in Figure \ref{fig:model}. The intention is to endow the model with this prior knowledge on sentence alignment between the source and target, and thus more easily attend to the appropriate sentences based on sentence locality. More specifically, on the source side, we do not use any sentence separator tokens; on the target side, we start each sentence with a sentence separator. During inference we initialize the output hypothesis with empty $\langle s \rangle$ sentence separator tokens, where the number of $\langle s \rangle$ equals to the number of source sentences, which is equal to the number of target sentences to be generated. These $\langle s \rangle$ tokens serve as sentence anchor points, and have sentence-positional information. Figure \ref{fig:model} visualizes the model.

In this work we increased the context window size to cover multiple sentences or a short document. Note that there is only a limit on the maximum number of tokens in the entire sequence; there is no limit on the length of a single sentence, or the total number of sentences in the sequence.

\section{Experiments}

We experiment with the WMT'19 English$\rightarrow$German document-level translation task \cite{barrault-acl-2019}. The training dataset consists of parallel document-level data (Europarl, Rapid, News-Commentary) and parallel sentence-level data (WikiTitles, Common Crawl, Paracrawl). The test set is newstest2019. The document-level portion contains 68.4k parallel documents, or a total of 7.7M parallel sentences; while the sentence-level portion has 19.9M parallel sentences. We generated a vocabulary of 32k subwords from the training data using the SentencePiece tokenizer \cite{kudo2018sentencepiece}.

The Big BIRD model is as described in Section \ref{sec:bigbird}, and the baseline Insertion Transformer model has exactly the same configurations except without sentence-positional embeddings. To be explicit, our baseline Insertion Transformer model is also given the prior knowledge of number of source sentences in the document. The target canvas is initialized target with $\langle s \rangle$ sentence separator tokens, where the number of $\langle s \rangle$ tokens is equal to the number of sentences in the document. All our models follow the same architecture as the Transformer Base model in \cite{vaswani-nips-2017}, and a context window of 1536 tokens during training (determined based on the longest document in the test set). All models were trained with the SM3 optimizer \cite{anil-arxiv-2019} with momentum 0.9, learning rate 0.1, and a quadratic learning rate warm-up schedule with 10k warm-up steps. The learning rate were chosen after some preliminary comparison runs between Adam and SM3. We opted to use the SM3 optimizer over Adam due to its more memory efficient properties, thus allowing us to use larger minibatches. Training was around 800k steps at batch size 512.

During training, each batch consists of 256 sub-documents and 256 sentences. Sub-documents are continuous sentences dynamically sampled from a document. The lengths of sub-documents are uniformly sampled in (0, 1536] tokens. The number of sampled sub-documents from each document is 1/10 of the number of sentences in the full document. Sentences directly come from sentence-level data. This 1:1 mixing of sub-documents and sentences results in training examples of vastly different lengths and therefore many masked positions, and we plan to improve it in the future by packing multiple sentences into one example.

We report sacreBLEU \cite{post-wmt-2018} scores of the two models in Table \ref{tab:results}. Our Big BIRD model outperforms the Insertion Transformer model by +4.3 BLEU.

\begin{table}[t]
\begin{center}
\begin{tabular}{lc}
\toprule
\bfseries{Model} & \bfseries{BLEU} \\ 
\midrule
Insertion Transformer & 25.3 \\
\midrule
Big BIRD & 29.6 \\
\bottomrule
\end{tabular}
\end{center}
\caption{WMT19 English$\rightarrow$German Document-Level Translation.}
\vspace{-0.6cm}
\label{tab:results}
\end{table}

\begin{table*}[t]
\begin{center}
\begin{tabular}{p{\textwidth}}
\toprule
\textbf{Source:} \\
(...) Chelsea faces Videoton in the UEFA Europa Leaguge at 3 p.m. on Thursday in London. \\
\midrule
\textbf{Target:} \\
(...) Chelsea trifft in der UEFA Europa League am Donnerstag um 15 Uhr in London auf Videoton. \\
\midrule
\textbf{Insertion Transformer:} \\
(...) Chelsea Gesichter am Donnerstag um 15.00 Uhr in London. Chelsea Gesichter Videoton in der UEFA Europa Leaguge. \\
\textbf{Translation:} (Google Translate) \\
Chelsea faces on Thursday at 15.00 in London. Chelsea faces Videoton in UEFA Europa Leaguge. \\
\midrule
\textbf{Big BIRD:} \\
(...) Chelsea sieht am Donnerstag um 15.00 Uhr in London Videoton in der UEFA Europa Leaguge. \\
\textbf{Translation:} (Google Translate) \\
Chelsea sees Videoton in UEFA Europa League on Thursday at 15.00 in London. \\
\bottomrule
\end{tabular}
\end{center}
\caption{An example where the Insertion Transformer gets confused with sentence alignment: it maps one sentence from the source into two sentences in the translation and loses semantic accuracy. When given sentence alignment explicitly, i.e. Big BIRD, it translates the sentence coherently.}
\label{tab:example}
\end{table*}

When we inspected the outputs more closely for the two models, we uncovered an interesting phenomenon. The Insertion Transformer, even though its target canvas is also initialized with the correct number of sentence $\langle s \rangle$ separators, struggles to align source and target sentences. For example, it can map two sources sentences into one sentence in the target, or vice versa. This is not always bad, as long as it captures the semantics accurately. However, there are cases when misalignment causes loss of coherency. Table \ref{tab:example} shows such an example where Big BIRD captures alignment better than the Insertion Transformer, and therefore its translation is more accurate and coherent.

\section{Conclusion}

In this paper, we presented Big BIRD, an adaptation of the Insertion Transformer to document-level translation. In addition to a large context window, Big BIRD also uses sentence-positional embeddings to directly capture sentence alignment between source and target documents. We show both quantitatively and qualitatively the promise of Big BIRD, with a +4.3 BLEU improvement over the baseline model and examples where Big BIRD achieves better translation quality via sentence alignment. We believe Big BIRD is a promising direction for document level understanding and generation.



\bibliography{acl2019}
\bibliographystyle{acl_natbib}



\end{document}